\documentclass[10pt,twocolumn,letterpaper]{article}
\pdfoutput=1
\usepackage{cvpr}
\usepackage{times}
\usepackage{epsfig}
\usepackage{graphicx}
\usepackage{amsmath}
\usepackage{amssymb}
\usepackage{stackengine}

\cvprfinalcopy 

\def\cvprPaperID{****} 

\ifcvprfinal\fi
\begin{document}

\title{Rethinking the Inception Architecture for Computer Vision}

\author{Christian Szegedy\\
Google Inc.\\
{\tt\small szegedy@google.com}
\and
Vincent Vanhoucke\\
{\tt\small vanhoucke@google.com}
\and
Sergey Ioffe\\
{\tt\small sioffe@google.com}
\and
Jonathon Shlens\\
{\tt\small shlens@google.com}
\and
Zbigniew Wojna\\
University College London\\
{\tt\small zbigniewwojna@gmail.com}
}

\maketitle

\begin{abstract}
Convolutional networks are at the core of most state-of-the-art
computer vision solutions for a wide variety of tasks.
Since 2014 very deep convolutional networks started to become
mainstream, yielding substantial gains in various benchmarks.
Although increased model size and computational cost tend to
translate to immediate quality gains for most tasks (as long as enough
labeled data is provided for training), computational efficiency and
low parameter count are still enabling factors for various use
cases such as mobile vision and big-data scenarios.
Here we are exploring ways to scale up networks in ways that aim at
utilizing the added computation as efficiently as possible by
suitably factorized convolutions and aggressive regularization.
We benchmark our methods on the ILSVRC 2012 classification challenge
validation set demonstrate substantial gains over the state of the art:
$21.2\%$ top-$1$ and $5.6\%$ top-$5$ error for {\it single frame}
evaluation using a network with a computational cost of $5$ billion
multiply-adds per inference and with using less than 25 million
parameters. With an ensemble of $4$ models and multi-crop
evaluation, we report $3.5\%$ top-$5$ error and $17.3\%$
top-$1$ error.

\end{abstract}

\section{Introduction}

Since the 2012 ImageNet competition~\cite{russakovsky2014imagenet}
winning entry by Krizhevsky et al~\cite{krizhevsky2012imagenet},
their network ``AlexNet'' has been successfully applied to a larger variety of
computer vision tasks, for example to object-detection~\cite{girshick2014rcnn},
segmentation~\cite{long2015fully}, human pose estimation~\cite{toshev2014deeppose},
video classification~\cite{karpathy2014large}, object
tracking~\cite{wang2013learning}, and superresolution~\cite{dong2014learning}.

These successes spurred a new line of research that focused on finding
higher performing convolutional neural networks. Starting in 2014, the
quality of network architectures significantly improved
by utilizing deeper and wider networks. VGGNet~\cite{simonyan2014very} and
GoogLeNet~\cite{szegedy2015going}
yielded similarly high performance in the 2014 ILSVRC~\cite{russakovsky2014imagenet}
classification challenge. One interesting observation was that gains in the
classification performance tend to transfer to significant quality gains in a
wide variety of application domains. This means that architectural improvements
in deep convolutional architecture can be utilized for improving performance for
most other computer vision tasks that are increasingly reliant on high quality,
learned visual features.
Also, improvements in the network quality resulted in new application
domains for convolutional networks in cases where AlexNet features
could not compete with hand engineered, crafted solutions,
e.g. proposal generation in detection\cite{erhan2014scalable}.

Although VGGNet \cite{simonyan2014very} has the compelling feature
of architectural simplicity, this comes at a high cost: evaluating the
network requires a lot of computation. On the other hand, the Inception
architecture of GoogLeNet \cite{szegedy2015going} was also designed to
perform well even under strict constraints on memory and computational budget.
For example, GoogleNet employed only 5 million parameters,
which represented a $12\times$ reduction with respect to its predecessor
AlexNet, which used $60$ million parameters.
Furthermore, VGGNet employed about 3x more parameters than AlexNet.

The computational cost of Inception is also much lower than VGGNet or its
higher performing successors~\cite{he2015delving}. This has made it feasible to
utilize Inception networks in big-data scenarios\cite{schroff2015facenet},
\cite{movshovitz2015ontological}, where huge amount of data needed to be
processed at reasonable cost or scenarios where memory or
computational capacity is inherently limited, for example in mobile vision
settings.
It is certainly possible to mitigate parts of these issues by applying
specialized solutions to target memory use~\cite{chen2015compressing},
\cite{psichogios1993svd}
or by optimizing the execution of certain operations via computational
tricks~\cite{lavin2015fast}. However, these methods add extra complexity.
Furthermore, these methods could be applied to optimize the
Inception architecture
as well, widening the efficiency gap again.

Still, the complexity of the Inception architecture makes
it more difficult to make changes to the network. If the architecture is
scaled up naively, large parts of the computational gains can be immediately
lost. Also, \cite{szegedy2015going} does not provide a clear description
about the contributing factors that lead to the various design decisions
of the GoogLeNet architecture. This makes it much harder to adapt it to new
use-cases while maintaining its efficiency. For example, if it is deemed
necessary to increase the capacity of some Inception-style model, the simple
transformation of just doubling the number of all filter bank sizes
will lead to a 4x increase in both computational cost and
number of parameters. This might prove prohibitive or unreasonable
in a lot of practical scenarios, especially if the associated gains
are modest. In this paper, we start with describing a few general
principles and optimization ideas that that proved to be useful for scaling up
convolution networks in efficient ways. Although our principles
are not limited to Inception-type networks, they are easier to observe
in that context as the generic structure of the Inception style building
blocks is flexible enough to incorporate those constraints naturally.
This is enabled by the generous use of dimensional reduction and
parallel structures of the Inception modules which allows for mitigating
the impact of structural changes on nearby components.
Still, one needs to be cautious about doing so, as some guiding principles
should be observed to maintain high quality of the models.

\section{General Design Principles}
\label{principles}
Here we will describe a few design principles based on large-scale experimentation with various architectural choices with convolutional networks. At this point, the utility of the principles below are speculative and additional future experimental evidence will be necessary to assess their accuracy and domain of validity. Still, grave deviations from these principles tended to result in deterioration in the quality of the networks and fixing situations where those deviations were detected resulted in improved architectures in general.

\begin{enumerate}
  \item Avoid representational bottlenecks, especially early in the network. Feed-forward networks can be represented by an acyclic graph from the input layer(s) to the classifier or regressor. This defines a clear direction for the information flow. For any cut separating the inputs from the outputs, one can access the amount of information passing though the cut. One should avoid bottlenecks with extreme compression. In general the representation size should gently decrease from the inputs to the outputs before reaching the final representation used for the task at hand. Theoretically, information content can not be assessed merely by the dimensionality of the representation as it discards important factors like correlation structure; the dimensionality merely provides a rough estimate of information content. \label{nobottlenecks}
  \item Higher dimensional representations are easier to process locally within a network. Increasing the activations per tile in a convolutional network allows for more disentangled features. The resulting networks will train faster. \label{highdim}
  \item Spatial aggregation can be done over lower dimensional embeddings without much or any loss in representational power. For example, before performing a more spread out (e.g. $3\times 3$) convolution, one can reduce the dimension of the input representation before the spatial aggregation without expecting serious adverse effects. We hypothesize that the reason for that is the strong correlation between adjacent unit results in much less loss of information during dimension reduction, if the outputs are used in a spatial aggregation context. Given that these signals should be easily compressible, the dimension reduction even promotes faster learning. \label{lowdim}
\item Balance the width and depth of the network. Optimal performance of the network can be reached by balancing the number of filters per stage and the depth of the network. Increasing both the width and the depth of the network can contribute to higher quality networks. However, the optimal improvement for a constant amount of computation can be reached if both are increased in parallel. The computational budget should therefore be distributed in a balanced way between the depth and width of the network. \label{balance}
\end{enumerate}

Although these principles might make sense, it is not straightforward to use
them to improve the quality of networks out of box. The idea is to use them
judiciously in ambiguous situations only.

\section{Factorizing Convolutions with Large Filter Size}

Much of the original gains of the  GoogLeNet network~\cite{szegedy2015going}
arise from a very generous use of dimension reduction. This can be viewed
as a special case of factorizing convolutions in a computationally efficient
manner. Consider for example the case of a $1\times 1$ convolutional layer
followed by a $3\times 3$ convolutional layer.
In a vision network, it is expected that the outputs
of near-by activations are highly correlated. Therefore,
we can expect that their activations can be reduced before
aggregation and that this should result in
similarly expressive local representations.

Here we explore other
ways of factorizing convolutions in various settings, especially in order to
increase the computational efficiency of the solution. Since Inception networks
are fully convolutional, each weight corresponds to one multiplication per
activation. Therefore, any reduction in computational cost results in reduced
number of parameters. This means that with suitable factorization, we can end up
with more disentangled parameters and therefore with faster training.
Also, we can use the computational and memory savings to increase the
filter-bank sizes of our network while maintaining our ability to train
each model replica on a single computer.

\subsection{Factorization into smaller convolutions}
\label{factorizing}
\begin{figure}
\centering
\includegraphics[width=\linewidth]{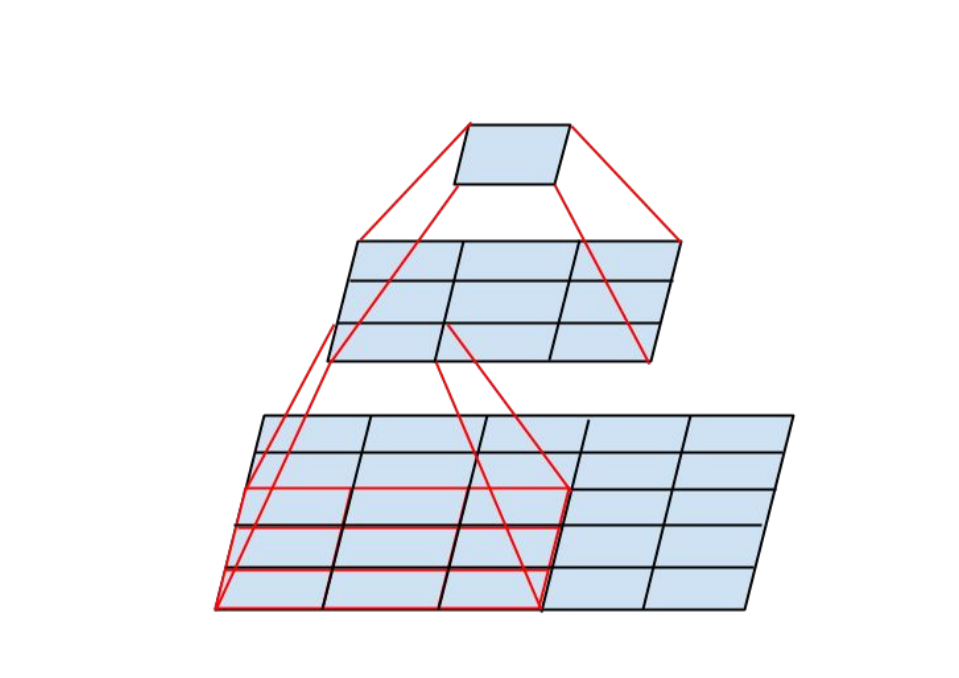}
\caption{Mini-network replacing the $5\times 5$ convolutions.}
\label{fig:double3}
\end{figure}

Convolutions with larger spatial filters (e.g. $5\times 5$ or
$7\times 7$) tend to be disproportionally expensive in terms of computation.
For example, a $5\times 5$ convolution
with $n$ filters over a grid with $m$ filters is 25/9 ~= 2.78 times
more computationally expensive
than a $3\times 3$ convolution with the same number of filters. Of course, a
$5\times 5$
filter can capture dependencies between signals between activations of units
further away in the earlier layers, so a reduction of the geometric size of the
filters comes at a large cost of expressiveness. However,
we can ask whether a $5\times 5$ convolution could be
replaced by a multi-layer network with less parameters with the same input
size and output depth. If we zoom into the computation graph of the
$5\times 5$ convolution, we see that each output looks like a small
fully-connected network sliding over $5\times 5$ tiles over its input
(see Figure~\ref{fig:double3}).
Since we are constructing a vision network, it seems natural to
exploit translation invariance again and replace the fully connected component
by a two layer convolutional architecture: the first layer is a $3\times 3$
convolution, the second is a fully connected layer on top of the $3\times 3$
output grid of the first layer (see Figure~\ref{fig:double3}).
Sliding this small network over the input activation grid boils down to
replacing the $5\times 5$ convolution with two layers of $3\times 3$
convolution (compare Figure~\ref{fig:inceptionv1} with \ref{fig:inceptionv2}).

\begin{figure}
\centering
\includegraphics[width=\linewidth]{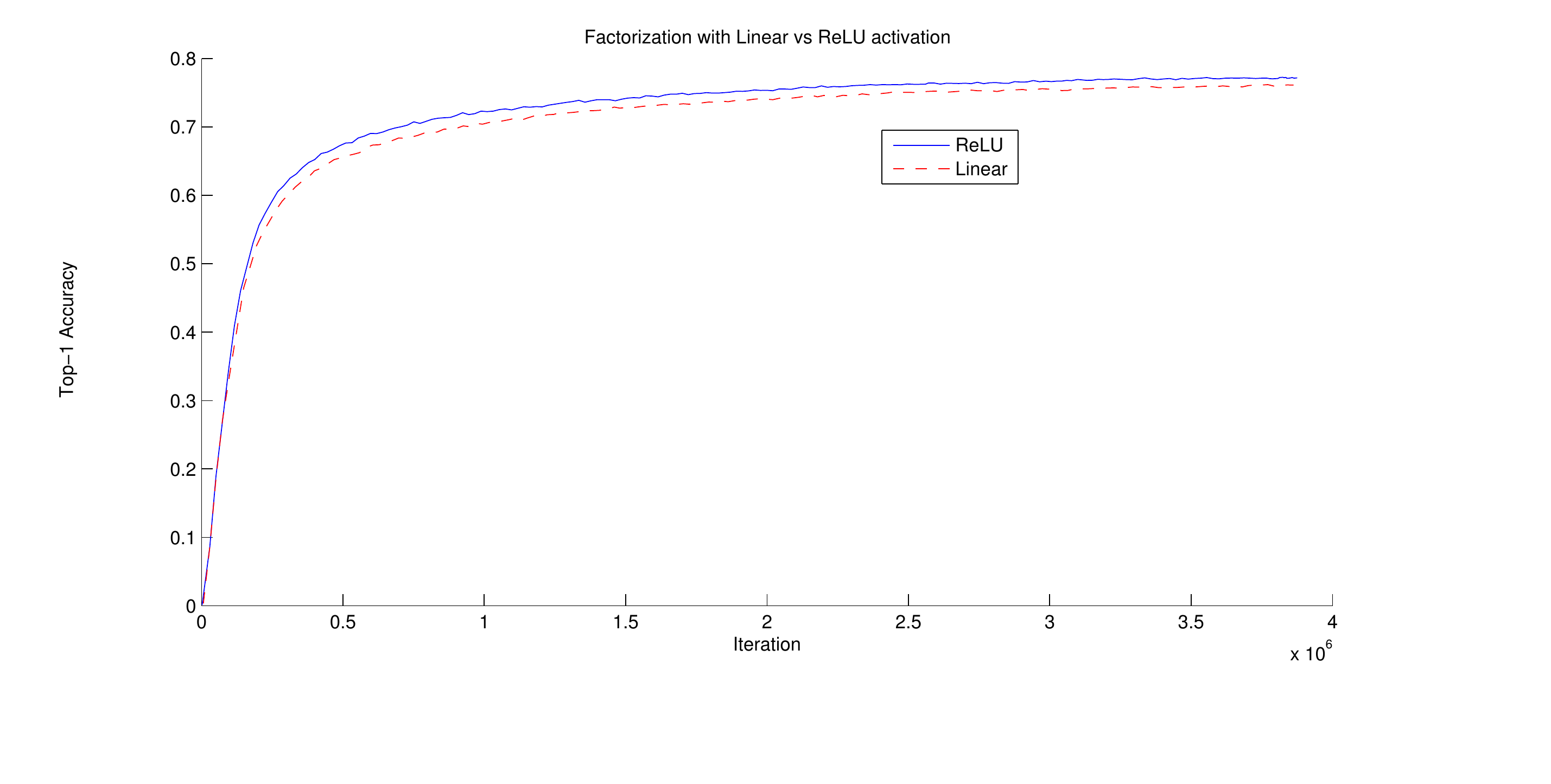}
\caption{One of several control experiments between two Inception models,
  one of them uses factorization into linear + ReLU layers, the other uses
  two ReLU layers. After $3.86$ million operations, the former settles at
  $76.2\%$, while the latter reaches $77.2\%$ top-1 Accuracy on the
  validation set.}
\label{fig:factorization_comparison}
\end{figure}

This setup clearly reduces the parameter count
by sharing the weights between adjacent tiles.
To analyze the expected computational cost savings,
we will make a few simplifying assumptions that apply for the typical
situations: We can assume that $n=\alpha m$, that is that we want to change the
number of activations/unit by a constant alpha factor. Since the $5\times 5$
convolution is aggregating, $\alpha$ is typically slightly larger
than one (around 1.5 in the case of GoogLeNet). Having a two layer replacement
for the $5\times 5$ layer, it seems reasonable to reach this expansion in two
steps: increasing the number of filters by $\sqrt{\alpha}$ in both steps.
In order to simplify our estimate by choosing $\alpha=1$ (no expansion),
If we would naivly slide a network without reusing the computation between
neighboring grid tiles, we would increase the computational cost.
sliding this network can be represented by two $3\times 3$ convolutional layers
which reuses the activations between adjacent tiles.
This way, we end up with a net $\frac{9+9}{25}\times$ reduction
of computation, resulting in a relative gain of $28\%$ by this factorization.
The exact same saving holds for the parameter count as each parameter
is used exactly once in the computation of the activation of each unit.
Still, this setup raises two general questions: Does this replacement result in
any loss of expressiveness? If our main goal is to factorize the linear part of
the computation, would it not suggest to keep linear activations in
the first layer? We have ran several control experiments
(for example see figure~\ref{fig:factorization_comparison}) and using
linear activation was always inferior to using rectified linear units in
all stages of the factorization.
We attribute this gain to the enhanced space of variations that the network
can learn especially if we batch-normalize~\cite{ioffe2015batch} the
output activations. One can see similar effects when using linear activations
for the dimension reduction components.

\subsection{Spatial Factorization into Asymmetric Convolutions}
\begin{figure}
\centering
\includegraphics[width=\linewidth]{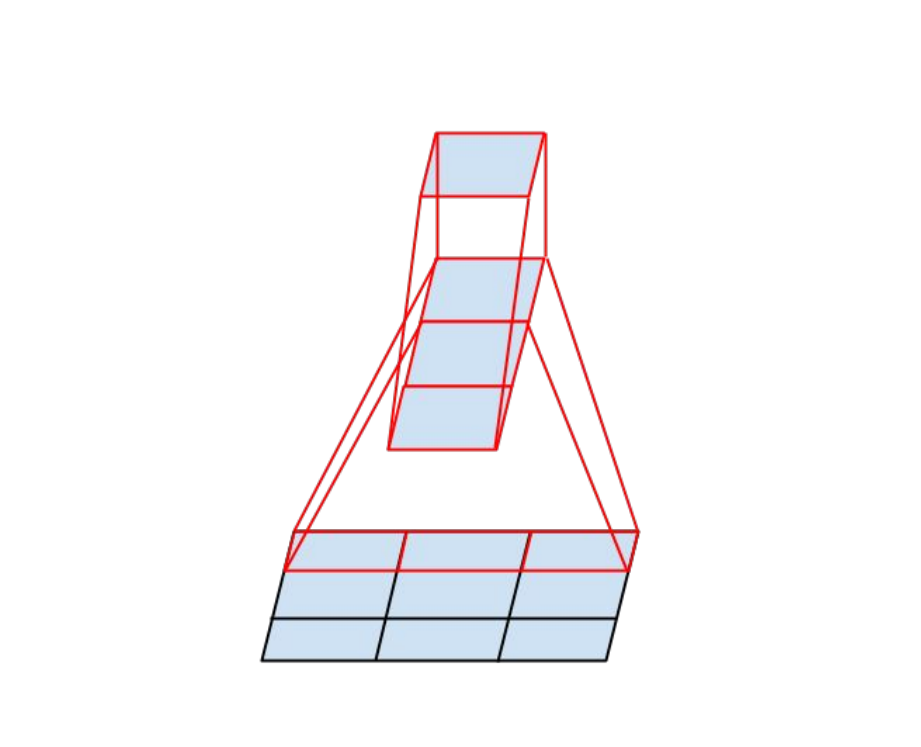}
\caption{Mini-network replacing the $3\times 3$ convolutions.
  The lower layer of this network consists of a $3\times 1$ convolution with
  $3$ output units.}
\label{fig:double31}
\end{figure}

\begin{figure}
\centering
\includegraphics[width=\linewidth]{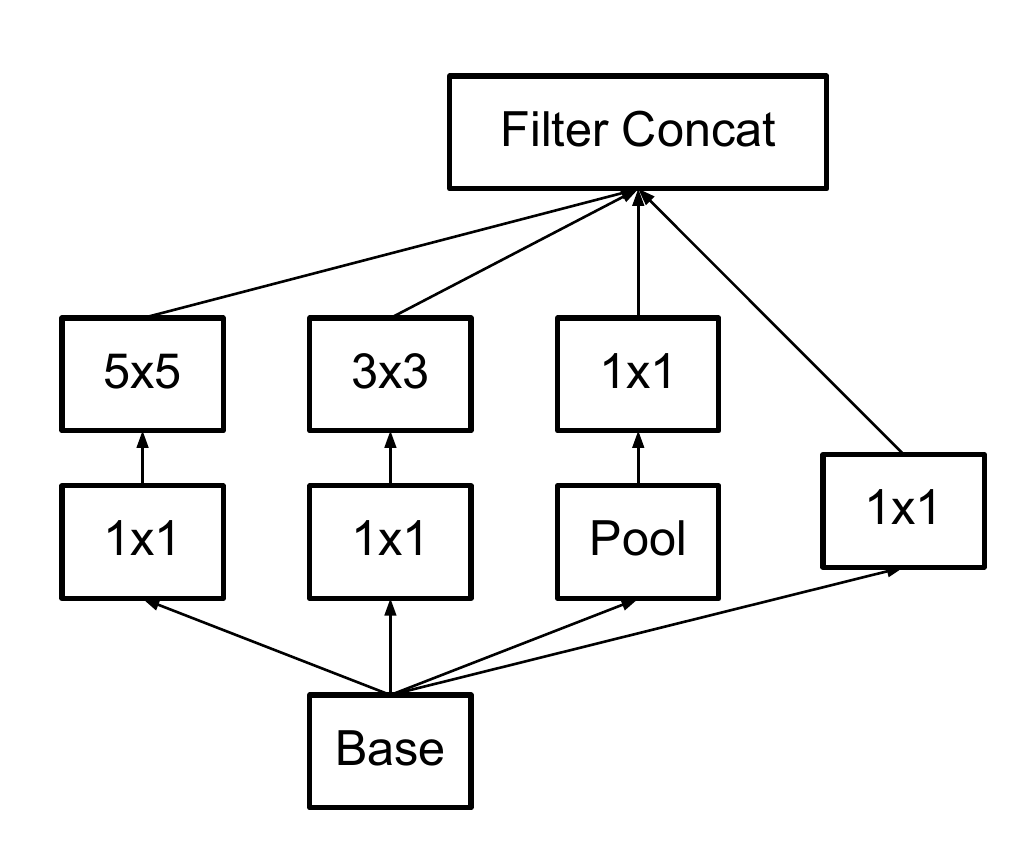}
\caption{Original Inception module as described in~\cite{szegedy2015going}.}
\label{fig:inceptionv1}
\end{figure}

\begin{figure}
\centering
\includegraphics[width=\linewidth]{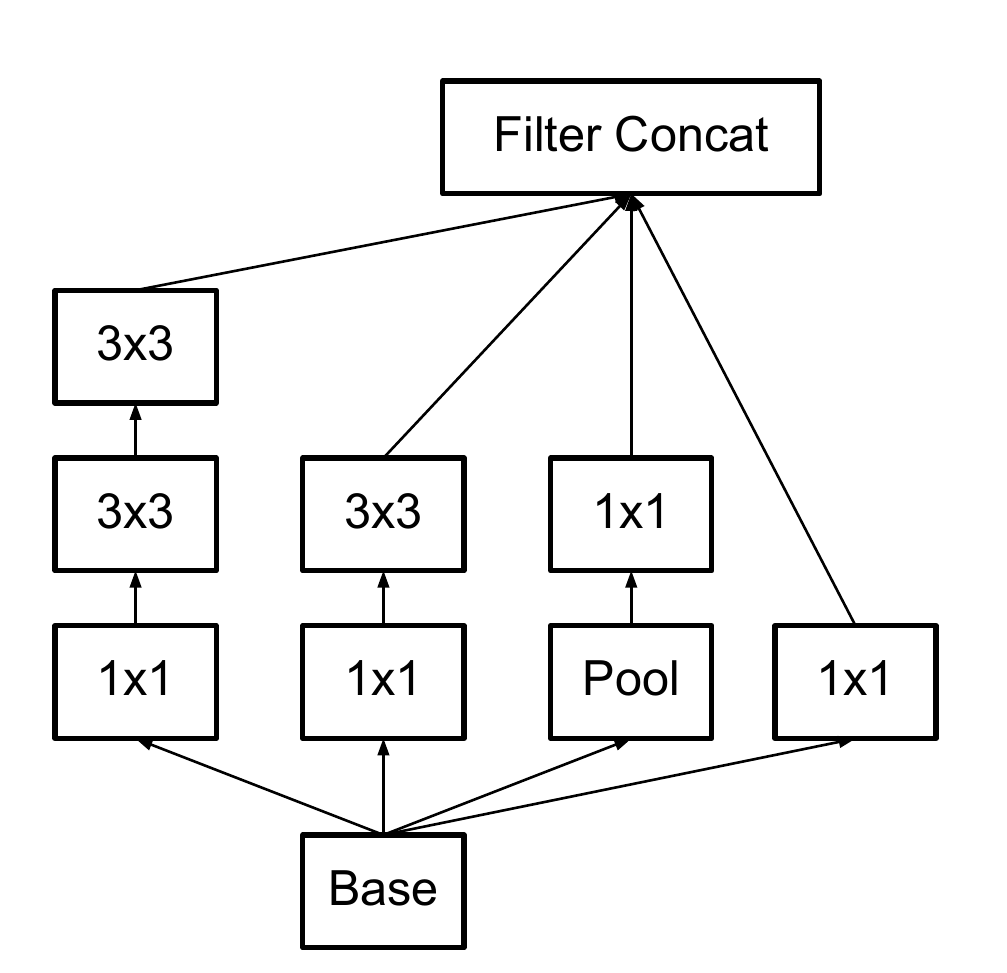}
\caption{Inception modules where each $5\times 5$ convolution is replaced by
  two $3\times 3$ convolution, as suggested by principle~\ref{lowdim} of Section~\ref{principles}.}
\label{fig:inceptionv2}
\end{figure}

The above results suggest that convolutions with filters larger
$3\times 3$  a might
not be generally useful as they can always be reduced into a sequence of
$3\times 3$ convolutional layers.
Still we can ask the question whether one should factorize them into smaller,
for example $2\times 2$ convolutions.
However, it turns out that one can do even better than $2\times 2$
by using asymmetric convolutions, e.g. $n\times 1$.
For example using a $3\times 1$ convolution followed by a $1\times 3$
convolution is equivalent to sliding a two layer network with the same
receptive field as in a $3\times 3$ convolution (see figure~\ref{fig:double31}).
Still the two-layer solution is $33\%$ cheaper for the same number of
output filters, if the number of input and output filters is equal.
By comparison, factorizing a $3\times 3$
convolution into a two $2\times 2$ convolution represents only a $11\%$ saving
of computation.
\begin{figure}
\centering
\includegraphics[width=\linewidth]{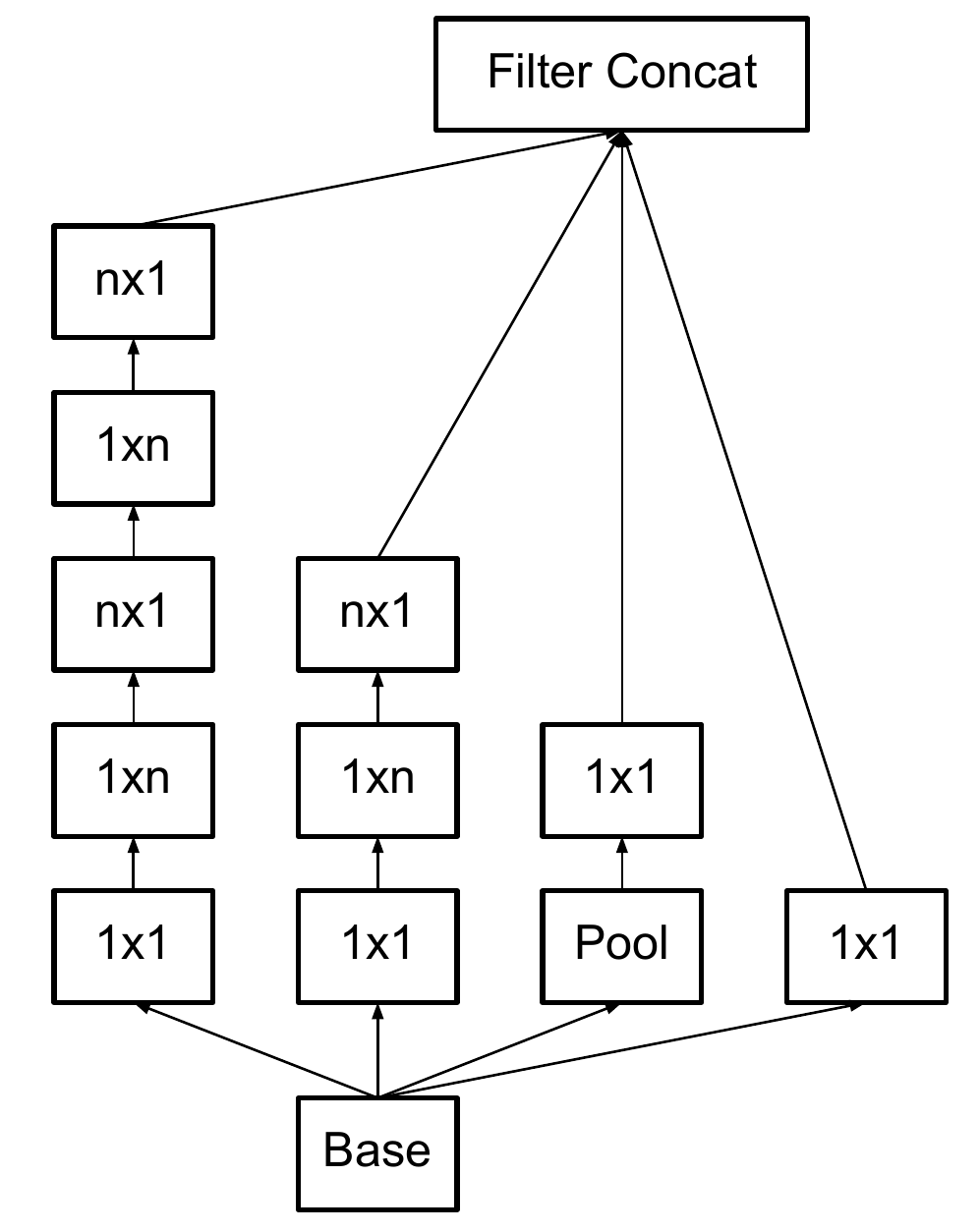}
\caption{Inception modules after the factorization of the $n\times n$
convolutions. In our proposed architecture, we chose $n=7$ for the
$17\times 17$ grid. (The filter sizes are picked using principle~\ref{lowdim})}.
\label{fig:inceptionv3}
\end{figure}
\begin{figure}
\centering
\includegraphics[width=\linewidth]{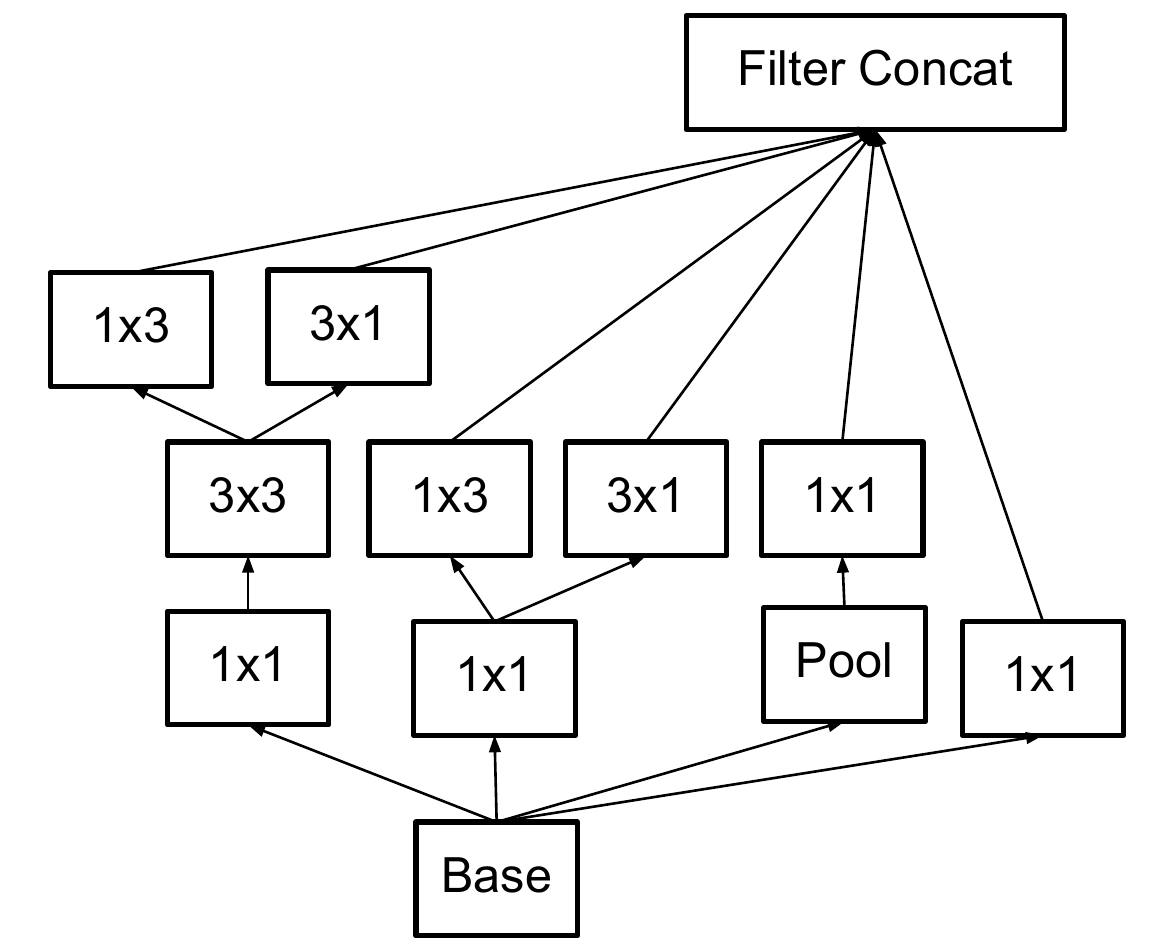}
\caption{Inception modules with expanded the filter bank outputs. This
  architecture is used on the coarsest ($8\times 8$) grids to promote
high dimensional representations, as suggested by principle~\ref{highdim} of Section~\ref{principles}.
We are using this solution only on the coarsest grid, since that is the place
where producing high dimensional sparse representation is the most critical
as the ratio of local processing (by $1\times 1$ convolutions) is increased
compared to the spatial aggregation.}
\label{fig:inceptionv4}
\end{figure}

In theory, we could go even further and argue that one can
replace any $n\times n$ convolution by a $1\times n$ convolution followed
by a $n\times 1$ convolution and the computational cost saving increases
dramatically as $n$ grows (see figure 6). In practice, we have found that employing this
factorization does not work well on early layers, but it gives very good results
on medium grid-sizes (On $m\times m$ feature maps, where $m$ ranges between $12$
and $20$). On that level, very good results can be achieved by using
$1\times 7$ convolutions followed by $7\times 1$ convolutions.

\section{Utility of Auxiliary Classifiers}

\cite{szegedy2015going} has introduced the notion of auxiliary classifiers
to improve the convergence of very deep networks. The original motivation was
to push useful gradients to the lower layers to make them immediately
useful and improve the convergence during training by combating the
vanishing gradient problem in very deep networks. Also Lee et al\cite{lee2014deeply}
argues that auxiliary classifiers promote more stable learning and better
convergence.
\begin{figure}
\centering
\includegraphics[width=\linewidth]{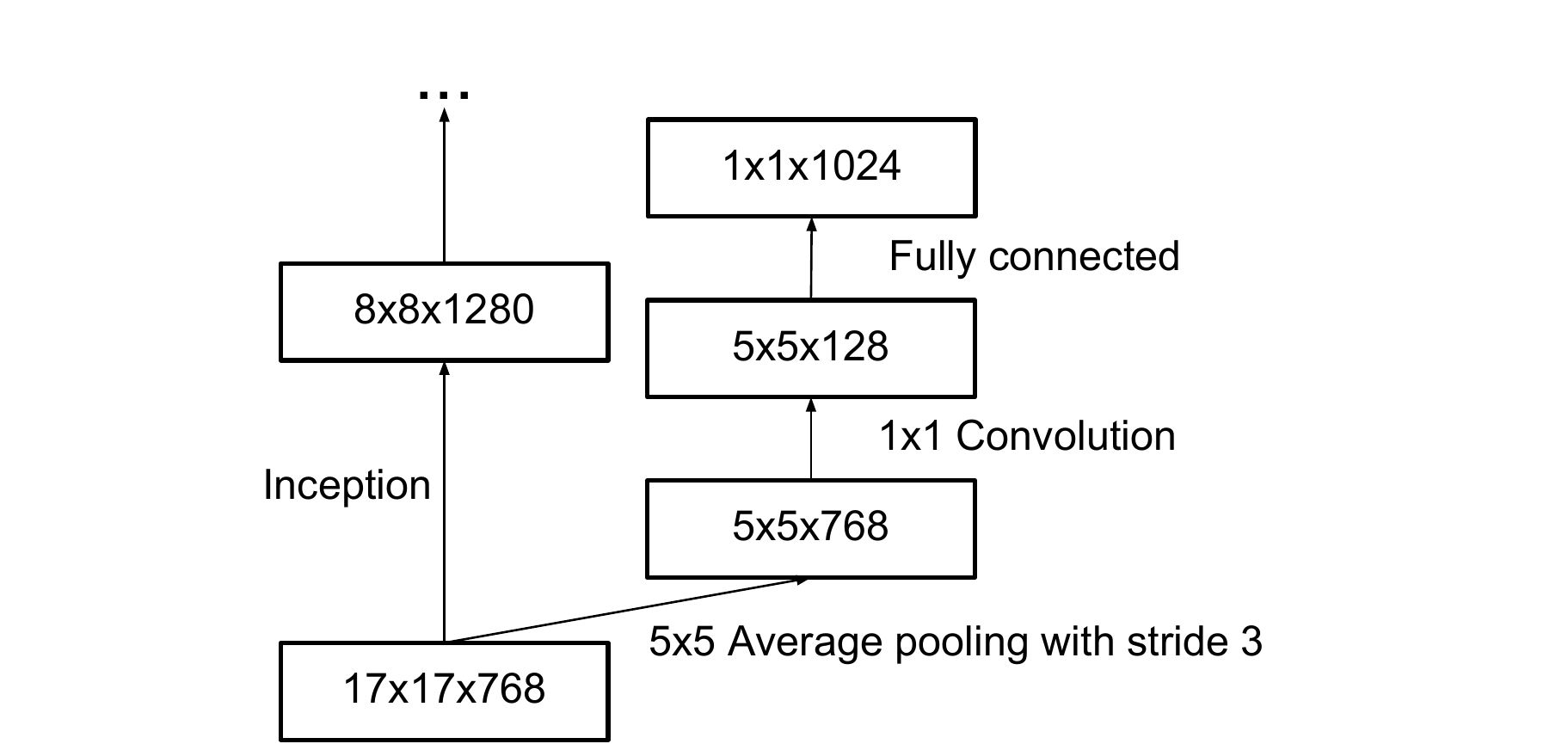}
\caption{Auxiliary classifier on top of the last $17\times 17$ layer.
Batch normalization\cite{ioffe2015batch} of the layers in the side head
results in a $0.4\%$ absolute gain in top-$1$ accuracy. The lower axis shows
the number of itertions performed, each with batch size $32$.}
\label{fig:sidehead}
\end{figure}
Interestingly, we found that auxiliary classifiers did not result in improved
convergence early in the training: the training progression of network with
and without side head looks virtually identical before both models reach
high accuracy. Near the end of training, 
the network with the auxiliary branches starts to
overtake the accuracy of the network without any auxiliary branch and
reaches a slightly higher plateau.

Also \cite{szegedy2015going} used two side-heads at different stages in the
network. The removal of the lower auxiliary branch did not have any
adverse effect on the final quality of the network. Together with the
earlier observation in the previous paragraph, this means that
original the hypothesis of \cite{szegedy2015going} that these branches
help evolving the low-level features is most likely misplaced.
Instead, we argue that the auxiliary classifiers act as regularizer.
This is supported by the fact that the main classifier of the
network  performs better if the side branch is batch-normalized~\cite{ioffe2015batch}
or has a dropout layer. This also gives a weak supporting evidence for the
conjecture that batch normalization acts as a regularizer.

\section{Efficient Grid Size Reduction}
\label{gridred}
\begin{figure}
\centering
\includegraphics[width=\linewidth]{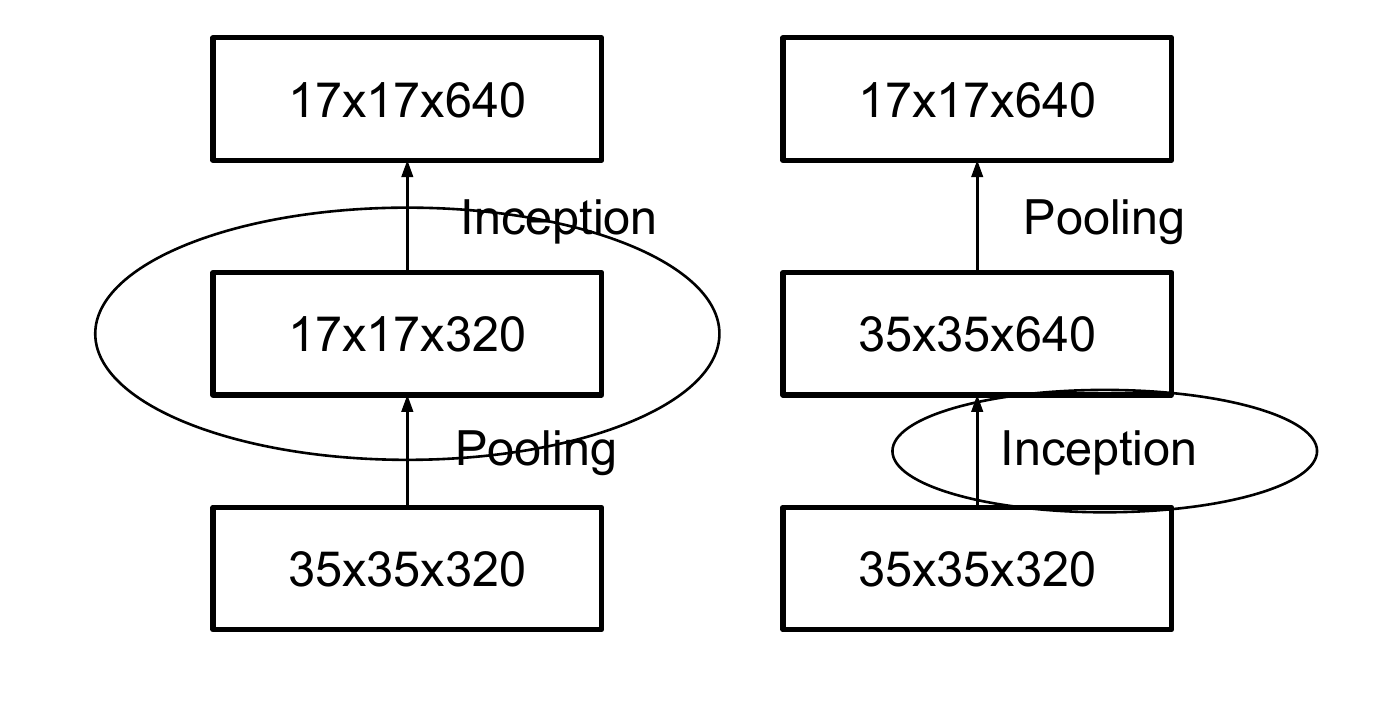}
\caption{Two alternative ways of reducing the grid size. The solution on the
  left violates the principle~\ref{nobottlenecks} of
  not introducing an representational bottleneck from Section~\ref{principles}.
  The version on the right is $3$ times more expensive computationally.
}
\label{fig:gridreduction}
\end{figure}
\begin{figure}
\centering
\includegraphics[width=\linewidth]{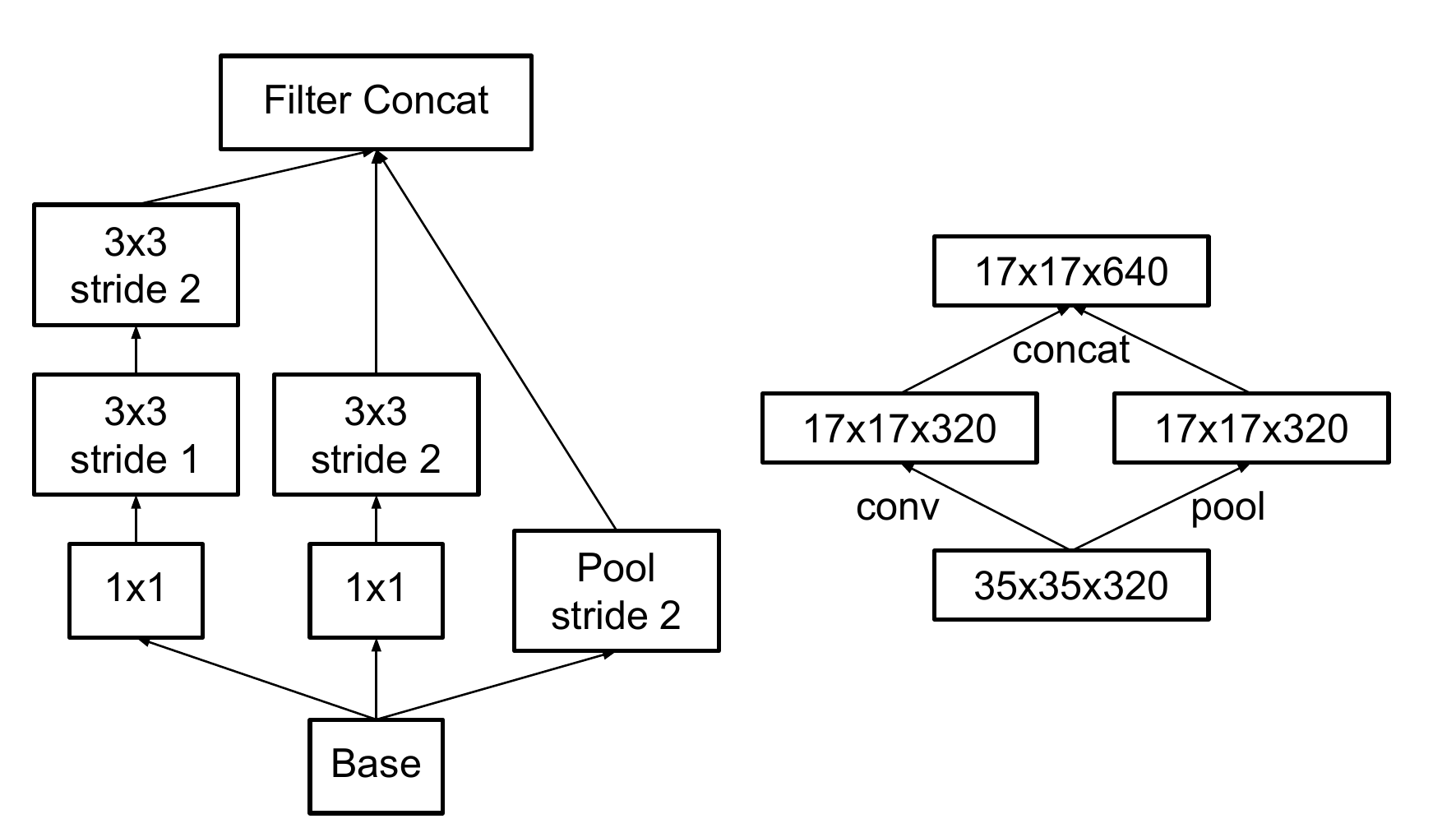}
\caption{Inception module that reduces the grid-size while expands the filter
  banks. It is both cheap and avoids the representational bottleneck as is
  suggested by principle~\ref{nobottlenecks}. 
  The diagram on the right represents the same solution but from the
  perspective of grid sizes rather than the operations.}
\label{fig:hybridreduction}
\end{figure}
Traditionally, convolutional networks used some pooling operation to decrease
the grid size of the feature maps. In order to avoid a representational
bottleneck, before applying maximum or average pooling the activation
dimension of the network filters is expanded.
For example, starting a $d\times d$ grid
with $k$ filters, if we would like to arrive at a
$\frac{d}{2}\times \frac{d}{2}$ grid with $2k$ filters,
we first need to compute a stride-1 convolution with $2k$
filters and then apply an additional pooling step. This means that the overall
computational cost is dominated by the expensive convolution on the larger grid
using $2d^2k^2$ operations. One possibility would be to switch to pooling with
convolution and therefore resulting in $2(\frac{d}{2})^2k^2$ reducing the
computational cost by a quarter. However, this creates a representational
bottlenecks as the overall dimensionality of the representation drops to
$(\frac{d}{2})^2k$ resulting in less expressive networks (see
Figure~\ref{fig:gridreduction}).
Instead of doing so, we suggest another variant the reduces the computational
cost even further while removing the representational bottleneck.
(see Figure~\ref{fig:hybridreduction}).
We can use two parallel stride 2 blocks: $P$ and $C$. $P$ is a pooling layer
(either average or maximum pooling) the activation, both of them are stride
$2$ the filter banks of which are concatenated as in
figure~\ref{fig:hybridreduction}.
\begin{table}
{\small
 \begin{center}
   \begin{tabular}[H]{|l|c|c|}
   \hline
   {\bf type} & \stackanchor{\bf patch size/stride}{or remarks} & {\bf input size} \\
   \hline\hline
   conv & $3{\times}3/2$ & $299{\times}299{\times}3$ \\
   \hline
    conv & $3{\times}3/1$ & $149{\times}149{\times}32$ \\
   \hline
   conv padded & $3{\times}3/1$ & $147{\times}147{\times}32$ \\
   \hline
   pool & $3{\times}3/2$ & $147{\times}147{\times}64$ \\
   \hline
   conv & $3{\times}3/1$ & $73{\times}73{\times}64$ \\
   \hline
   conv & $3{\times}3/2$ & $71{\times}71{\times}80$ \\
   \hline
   conv & $3{\times}3/1$ & $35{\times}35{\times}192$ \\
   \hline
   $3\times$Inception & As in figure~\ref{fig:inceptionv2} & $35{\times}35{\times}288$ \\
   \hline
   $5\times$Inception & As in figure~\ref{fig:inceptionv3} & $17{\times}17{\times}768$ \\
   \hline
   $2\times$Inception & As in figure~\ref{fig:inceptionv4} & $8{\times}8{\times}1280$ \\
   \hline
   pool & $8\times 8$ & $8\times 8\times 2048$ \\
   \hline
   linear & logits & $1\times 1\times 2048$ \\
   \hline
   softmax & classifier & $1\times 1\times 1000$ \\
   \hline
   \end{tabular}
 \end{center}
 }
\caption{The outline of the proposed network architecture.
  The output size of each module is the input size of the next one.
  We are using variations of reduction technique depicted
  Figure~\ref{fig:hybridreduction} to reduce the grid sizes between the
  Inception blocks whenever applicable.
  We have marked the convolution with $0$-padding,
  which is used to maintain the grid size. $0$-padding is also used
  inside those Inception modules that do not reduce the grid size.
  All other layers do not use padding. The various filter bank
  sizes are chosen to observe principle~\ref{balance} from
  Section~\ref{principles}.
}
\label{table:stem}
\end{table}

\section{Inception-v2}
\label{revisited}
Here we are connecting the dots from above and propose a new
architecture with improved performance on the ILSVRC 2012
classification benchmark.
The layout of our network is given in table~\ref{table:stem}.
Note that we have factorized the traditional $7\times 7$ convolution into
three $3\times 3$ convolutions based on the same ideas as described in
section~\ref{factorizing}.
For the Inception part of the network, we have $3$ traditional
inception modules at  the $35\times 35$ with $288$ filters each.
This is reduced to a $17\times 17$ grid with $768$ filters using the
grid reduction technique described in section \ref{gridred}. This is
is followed by $5$ instances of the factorized inception modules as
depicted in figure~\ref{fig:inceptionv2}. This is reduced to a $8\times 8\times 1280$
grid with the grid reduction technique depicted in figure \ref{fig:hybridreduction}.
At the coarsest $8\times 8$ level, we have two Inception modules as depicted
in figure~\ref{fig:inceptionv3}, with a concatenated output filter bank size of
2048 for each tile. The detailed structure of the
network, including the sizes of filter banks inside the Inception modules,
is given in the supplementary material, given in the {\tt model.txt} that is in
the tar-file of this submission. However, we have observed that
the quality of the network is relatively stable to variations
as long as the principles from Section~\ref{principles} are observed.
Although our network is $42$ layers deep, our computation cost is only
about $2.5$ higher than that of GoogLeNet and it is still much more efficient than
VGGNet.

\section{Model Regularization via Label Smoothing}
\label{smoothing}
Here we propose a mechanism to regularize the classifier layer by
estimating the marginalized effect of label-dropout during training.

For each training example $x$, our model computes the probability of each label
$k\in\{1\ldots K\}$: $p(k|x) = \frac{\exp(z_k)}{\sum_{i=1}^K \exp(z_i)}$. Here,
$z_i$ are the {\em logits} or unnormalized log-probabilities. Consider the
ground-truth distribution over labels $q(k|x)$ for this training example,
normalized so that $\sum_k q(k|x) = 1$. For brevity, let us omit the dependence
of $p$ and $q$ on example $x$. We define the loss for the example as the
cross entropy: $\ell = -\sum_{k=1}^K \log(p(k)) q(k)$. Minimizing this is
equivalent to maximizing the expected log-likelihood of a label, where the label
is selected according to its ground-truth distribution $q(k)$. Cross-entropy
loss is differentiable with respect to the logits $z_k$ and thus can be used for
gradient training of deep models. The gradient has a rather simple form:
$\frac{\partial\ell}{\partial z_k} = p(k) - q(k)$, which is bounded between $-1$
and $1$.

Consider the case of a single ground-truth label $y$, so that $q(y)=1$ and
$q(k)=0$ for all $k\neq y$. In this case, minimizing the cross entropy is
equivalent to maximizing the log-likelihood of the correct label. For a
particular example $x$ with label $y$, the log-likelihood is maximized for $q(k)
= \delta_{k,y}$, where $\delta_{k,y}$ is Dirac delta, which equals $1$ for $k=y$
and $0$ otherwise. This maximum is not achievable for finite $z_k$ but is
approached if $z_y\gg z_k$ for all $k\neq y$ -- that is, if the logit
corresponding to the ground-truth label is much great than all other
logits. This, however, can cause two problems. First, it may result in
over-fitting: if the model learns to assign full probability to the ground-truth
label for each training example, it is not guaranteed to generalize. Second, it
encourages the differences between the largest logit and all others to become
large, and this, combined with the bounded gradient
$\frac{\partial\ell}{\partial z_k}$, reduces the ability of the model to
adapt. Intuitively, this happens because the model becomes too confident about
its predictions.

We propose a mechanism for encouraging the model to be less confident. While
this may not be desired if the goal is to maximize the log-likelihood of
training labels, it does regularize the model and makes it more adaptable. The
method is very simple. Consider a distribution over labels $u(k)$, {\em
  independent of the training example $x$}, and a smoothing parameter $\epsilon$.
 For a training example with ground-truth label $y$, we
replace the label distribution $q(k|x)=\delta_{k,y}$ with
$$
q'(k|x) = (1-\epsilon) \delta_{k,y} + \epsilon u(k)
$$
which is a mixture of the original ground-truth distribution $q(k|x)$ and the
fixed distribution $u(k)$, with weights $1-\epsilon$ and $\epsilon$,
respectively. This can be seen as the distribution of the label $k$ obtained as
follows: first, set it to the ground-truth label $k=y$; then, with probability
$\epsilon$, replace $k$ with a sample drawn from the distribution $u(k)$. We
propose to use the prior distribution over labels as $u(k)$. In our experiments,
we used the uniform distribution $u(k) = 1/K$, so that
$$
q'(k) = (1-\epsilon) \delta_{k,y} + \frac{\epsilon}{K}.
$$
We refer to this change in ground-truth label distribution as {\em
  label-smoothing regularization}, or LSR.

Note that LSR achieves the desired goal of preventing the largest logit from
becoming much larger than all others. Indeed, if this were to happen, then a
single $q(k)$ would approach $1$ while all others would approach $0$. This would
result in a large cross-entropy with $q'(k)$ because, unlike
$q(k)=\delta_{k,y}$, all $q'(k)$ have a positive lower bound.

Another interpretation of LSR can be obtained by considering the cross entropy:
$$
H(q',p) = -\sum_{k=1}^K \log p(k) q'(k) = (1-\epsilon)H(q, p) + \epsilon H(u, p)
$$
Thus, LSR is equivalent to replacing a single cross-entropy loss $H(q,p)$ with a
pair of such losses $H(q,p)$ and $H(u,p)$. The second loss penalizes the
deviation of predicted label distribution $p$ from the prior $u$, with the 
relative weight $\frac{\epsilon}{1-\epsilon}$. Note that this deviation could be
equivalently captured by the KL divergence, since $H(u,p) = D_{KL}(u\|p) + H(u)$
and $H(u)$ is fixed. When $u$ is the uniform distribution, $H(u,p)$ is a measure
of how dissimilar the predicted distribution $p$ is to uniform, which could also be
measured (but not equivalently) by negative entropy $-H(p)$; we have not
experimented with this approach.

In our ImageNet experiments with $K=1000$ classes, we used $u(k) = 1/1000$ and
$\epsilon=0.1$. For ILSVRC 2012, we have found a consistent improvement of
about $0.2\%$ absolute both for top-$1$ error and the top-$5$ error
(cf. Table~\ref{results}).

\section{Training Methodology}
We have trained our networks with stochastic gradient utilizing the
TensorFlow~\cite{tensorflow2015-whitepaper} distributed machine learning system
using $50$ replicas running each on a NVidia Kepler GPU with batch size $32$
for $100$ epochs.
Our earlier experiments used momentum~\cite{icml2013_sutskever13} with a
decay of $0.9$, while our best models were achieved using RMSProp~\cite{rmsprop}
with decay of $0.9$ and $\epsilon=1.0$. We used a learning rate of $0.045$,
decayed every two epoch using an exponential rate of $0.94$.
In addition, gradient clipping \cite{pascanu2012difficulty} with threshold $2.0$
was found to be useful to stabilize the training. Model evaluations are
performed using a running average of the parameters computed over time.

\section{Performance on Lower Resolution Input}

A typical use-case of vision networks is for the the post-classification of
detection, for example in the Multibox~\cite{erhan2014scalable} context.
This includes
the analysis of a relative small patch of the image containing a single
object with some context. The tasks is to decide whether the center part
of the patch corresponds to some object and determine the class of the
object if it does. The challenge is that objects tend to be relatively
small and low-resolution. This raises the question of how to properly
deal with lower resolution input.

The common wisdom is that models employing higher resolution receptive
fields tend to result in significantly improved recognition performance.
However it is important to distinguish between the effect of the
increased resolution of the first layer receptive field and the
effects of larger model capacitance and computation.
If we just change the resolution of the input without further
adjustment to the model, then we end up using computationally much
cheaper models to solve more difficult tasks.
Of course, it is natural, that these solutions loose out already because of the
reduced computational effort. In order to make an accurate assessment,
the model needs to analyze vague hints in order to be able to
``hallucinate'' the fine details.
This is computationally costly. The question remains therefore: how
much does higher input resolution helps if the computational effort is
kept constant. One simple way to ensure constant effort is
to reduce the strides of the first two layer in the case of
lower resolution input, or by simply removing the first pooling layer of
the network.

For this purpose we have performed the following three experiments:
\begin{enumerate}
  \item $299\times 299$ receptive field with stride $2$ and maximum pooling
        after the first layer.
  \item $151\times 151$ receptive field with stride $1$ and maximum pooling
        after the first layer.
  \item $79\times 79$ receptive field with stride $1$ and {\bf without}
        pooling after the first layer.
\end{enumerate}
All three networks have almost identical computational cost. Although the third
network is slightly cheaper, the cost of the pooling layer is marginal
and (within $1\%$ of the total cost of the)network.
In each case, the networks were trained until convergence and their
quality was measured on the validation set of the ImageNet ILSVRC 2012
classification benchmark. The results can be seen in table~\ref{lowrescmp}.
Although the lower-resolution networks take longer to train,
the quality of the final result is quite close to that of their
higher resolution counterparts.

However, if one would just naively reduce the network size according to the
input resolution, then network would perform much more poorly. However this
would an unfair comparison as we would are comparing a 16 times cheaper model on
a more difficult task.

Also these results of table~\ref{lowrescmp} suggest, one might consider using
dedicated high-cost low resolution networks for smaller objects in the
R-CNN~\cite{girshick2014rcnn} context.

\begin{table}
{
 \begin{center}
   \begin{tabular}[H]{|l|l|}
   \hline
   {\bf Receptive Field Size} & {\bf Top-1 Accuracy (single frame)}\\
   \hline\hline
   $79\times 79$ & 75.2\% \\
   \hline
   $151\times 151$ & 76.4\% \\
   \hline
   $299\times 299$ & 76.6\% \\
   \hline
   \end{tabular}
 \end{center}
 }
\caption{Comparison of recognition performance when the size of the receptive
field varies, but the computational cost is constant.}
\label{lowrescmp}
\end{table}

\section{Experimental Results and Comparisons}
\begin{table}
{\small
 \begin{center}
   \begin{tabular}[H]{|l|c|c|c|}
   \hline
   {\bf Network} & \stackanchor{\bf Top-1}{Error} & \stackanchor{\bf Top-5}{Error} & \stackanchor{\bf Cost}{Bn Ops} \\
   \hline\hline
   GoogLeNet~\cite{szegedy2015going} & 29\% & 9.2\% & 1.5 \\
   \hline
   BN-GoogLeNet & 26.8\% & - & {\bf 1.5} \\
   \hline
   BN-Inception~\cite{ioffe2015batch} & 25.2\% & 7.8 & 2.0 \\
   \hline
   Inception-v2 & 23.4\% & - & 3.8 \\
   \hline
   \shortstack[l]{Inception-v2 \\RMSProp} & 23.1\% & 6.3 & 3.8 \\
   \hline
   \shortstack[l]{Inception-v2 \\ Label Smoothing} & 22.8\% & 6.1 & 3.8 \\
   \hline
   \shortstack[l]{Inception-v2 \\ Factorized $7\times 7$} & 21.6\% & 5.8 & 4.8 \\
   \hline
   \stackanchor{Inception-v2}{BN-auxiliary} & {\bf 21.2}\% & {\bf 5.6}\% & 4.8 \\
   \hline
   \end{tabular}
 \end{center}
 }
\caption{Single crop experimental results comparing the cumulative effects on
 the various contributing   factors. We compare our numbers with the best
 published single-crop inference for Ioffe at
 al~\cite{ioffe2015batch}. For the ``Inception-v2'' lines, the
 changes are cumulative and each subsequent line includes the new change in
 addition to the previous ones. The last line is referring to all the changes
 is what we refer to as ``Inception-v3'' below. Unfortunately, He et
 al~\cite{he2015delving} reports the only 10-crop evaluation results, but not
 single crop results, which is reported in the Table~\ref{resultsmulticrop}
 below.
}
 \label{results}
\end{table}
Table~\ref{results} shows the experimental results about the recognition
performance of our proposed architecture (Inception-v2) as described in
Section~\ref{revisited}. Each Inception-v2 line shows the result of the
cumulative changes including the highlighted new modification plus all the
earlier ones. Label Smoothing refers to method described in Section~\ref{smoothing}.
Factorized $7\times 7$ includes a change that factorizes the first $7\times 7$
convolutional layer into a sequence of $3\times 3$ convolutional layers.
BN-auxiliary refers to the version in which the fully connected layer
of the auxiliary classifier is also batch-normalized, not just the convolutions.
We are referring to the model in last row of Table~\ref{results} as Inception-v3
and evaluate its performance in the multi-crop and ensemble settings.

All our evaluations are done on the 48238 non-blacklisted examples on the
ILSVRC-2012 validation set, as suggested by ~\cite{russakovsky2014imagenet}.
We have evaluated all the 50000 examples as well and the results were roughly
0.1\% worse in top-5 error and around 0.2\% in top-1
error. In the upcoming version of this paper, we will verify our ensemble
result on the test set, but at the time of our last evaluation of
BN-Inception in spring~\cite{ioffe2015batch} indicates that the test and
validation set error tends to correlate very well.

\begin{table}
{\small
 \begin{center}
   \begin{tabular}[H]{|l|c|c|c|}
   \hline
   {\bf Network} &
   \stackanchor{\bf Crops}{\bf Evaluated} &
   \stackanchor{\bf Top-5}{\bf Error} &
   \stackanchor{\bf Top-1}{\bf Error} \\
   \hline\hline
   GoogLeNet~\cite{szegedy2015going} & 10 & - & 9.15\% \\
   \hline
   GoogLeNet~\cite{szegedy2015going} & 144 & - & 7.89\% \\
   \hline
   VGG~\cite{simonyan2014very} & - & 24.4\% & 6.8\% \\
   \hline
   BN-Inception~\cite{ioffe2015batch} & 144 & 22\% & 5.82\% \\
   \hline
   PReLU~\cite{he2015delving} & 10 & 24.27\% & 7.38\% \\
   \hline
   PReLU~\cite{he2015delving} & - & 21.59\% & 5.71\% \\
   \hline
   Inception-v3 & 12 & 19.47\% & 4.48\% \\
   \hline
   Inception-v3 & 144 & \bf{18.77\%} & \bf{4.2\%} \\
   \hline
   \end{tabular}
 \end{center}
 }
\caption{Single-model, multi-crop experimental results comparing the cumulative effects on
 the various contributing   factors. We compare our numbers with the best
 published single-model inference results on the ILSVRC 2012 classification
 benchmark.} \label{resultsmulticrop}
\end{table}

\begin{table}
{\small
 \begin{center}
   \begin{tabular}[H]{|l|c|c|c|c|}
   \hline
   {\bf Network} &
   \stackanchor{\bf Models}{\bf Evaluated} &
   \stackanchor{\bf Crops}{\bf Evaluated} &
   \stackanchor{\bf Top-1}{\bf Error} &
   \stackanchor{\bf Top-5}{\bf Error} \\
   \hline\hline
   VGGNet~\cite{simonyan2014very} & 2 & - & 23.7\% & 6.8\% \\
   \hline
   GoogLeNet~\cite{szegedy2015going} & 7 & 144 & - & 6.67\% \\
   \hline
   PReLU~\cite{he2015delving} & - & - & - & 4.94\% \\
   \hline
   BN-Inception~\cite{ioffe2015batch} & 6 & 144 & 20.1\% & 4.9\% \\
   \hline
   Inception-v3 & 4 & 144 & {\bf 17.2\%} & {\bf 3.58\%}$^*$ \\
   \hline
   \end{tabular}
 \end{center}
}
\caption{Ensemble evaluation results comparing multi-model, multi-crop
reported results. Our numbers are compared with the best  published
ensemble inference results on the ILSVRC 2012 classification benchmark.
$^*$All results, but the top-5 ensemble result reported are 
on the validation set. The ensemble yielded $3.46\%$ top-5 error on the
validation set.}
\label{resultsensemble}
\end{table}




\section{Conclusions}

We have provided several design principles to scale up convolutional networks
and studied them in the context of the Inception architecture. This
guidance can lead to high performance vision networks that have a relatively
modest computation cost compared to simpler, more monolithic architectures.
Our highest quality version of Inception-v3 reaches $21.2\%$,
top-$1$ and $5.6\%$ top-5 error for {\bf single crop} evaluation
on the ILSVR 2012 classification, setting a new state of the art.
This is achieved with relatively modest ($2.5\times$) increase in computational
cost compared to the network described in Ioffe et al \cite{ioffe2015batch}.
Still our solution uses much less computation than the best published
results based on denser networks: our model outperforms
the results of He et al \cite{he2015delving} -- cutting the top-$5$ (top-$1$)
error by $25\%$ ($14\%$) relative, respectively -- while
being six times cheaper computationally and using at least five times less
parameters (estimated). Our ensemble of four Inception-v3 models reaches
$3.5\%$ with multi-crop evaluation reaches $3.5\%$ top-$5$ error which 
represents an over $25\%$ reduction to the best published results and
is almost half of the error of ILSVRC 2014 winining GoogLeNet ensemble.

We have also demonstrated that high quality results can be reached with
receptive field resolution as low as $79\times 79$. This might prove
to be helpful in systems for detecting relatively small objects.
We have studied how factorizing convolutions and aggressive dimension
reductions inside neural network can result in networks with relatively
low computational cost while maintaining high quality.
The combination of lower parameter count and additional
regularization with batch-normalized auxiliary classifiers and
label-smoothing allows for training high quality networks on relatively
modest sized training sets.


{\small
\bibliographystyle{ieee}
\bibliography{references}
}

\end{document}